\documentclass{article}
\usepackage{spconf,amsmath,graphicx}

\usepackage{multirow}
\usepackage{amssymb}
\usepackage{lipsum}
\usepackage{enumitem}

\title{Personalization of CTC-based End-to-End Speech Recognition Using Pronunciation-Driven Subword Tokenization}
\name{
\begin{tabular}{@{}c@{}}
Zhihong Lei, Ernest Pusateri, Shiyi Han, Leo Liu, Mingbin Xu, Tim Ng, \\
Ruchir Travadi, Youyuan Zhang, Mirko Hannemann, Man-Hung Siu, Zhen Huang
\end{tabular}
}

\address{Apple}

\begin{document}
%
\maketitle
\begin{abstract}
Recent advances in deep learning and automatic speech recognition have improved the accuracy of end-to-end speech recognition systems, but recognition of personal content such as contact names remains a challenge. In this work, we describe our personalization solution for an end-to-end speech recognition system based on connectionist temporal classification. Building on previous work, we present a novel method for generating additional subword tokenizations for personal entities from their pronunciations.  We show that using this technique in combination with two established techniques, contextual biasing and wordpiece prior normalization, we are able to achieve personal named entity accuracy on par with a competitive hybrid system.

\end{abstract}
\section{Introduction}


The successful recognition of personal named entities, including a user's contacts, favorite music, and apps, is crucial for automatic speech recognition (ASR) applications such as voice assistants on mobile devices. However, recognition of personal named entities is challenging, particularly for end-to-end (E2E) ASR, because named entities often contain rare words, and many are pronounced or spelled irregularly. To compensate, ASR systems are frequently adapted to particular users, a process often called personalization.

In a hybrid ASR system, personalization is relatively straightforward. This type of system usually consists of an acoustic model (AM), a language model (LM), and a pronunciation lexicon. Assuming the AM is sufficiently trained on acoustic units that cover rare personal named entities, only the lexicon and LM need to be personalized. A typical solution in hybrid ASR systems is to introduce a contextual language model which is trained on personal named entities and represented by a finite state transducer (FST). During recognition, the contextual LM is combined with the general language model through on-the-fly FST operations to achieve high accuracy for personal named entities (e.g. \cite{aleksic2015improved}.)

In contrast to hybrid ASR systems, personalization of E2E ASR systems is quite challenging. E2E ASR systems use a single neural network to jointly model acoustics and language. These systems are trained on paired audio and text data and often do not perform well on entities that appear infrequently or appear with different pronunciations in their training data. Without separate LM and lexicon components, these models lack clear mechanisms for personalization.  Still, contextual biasing approaches have been applied to E2E systems with some success  \cite{zhao2019shallow}.  Wordpiece prior normalization, a technique which attempts to remove the LM component of the E2E model score so that it may be more effectively combined with an external LM, has provided an additional benefit \cite{sainath2021efficient}.   Further, Huang et al \cite{huang2020class} demonstrated accuracy improvements using a method to tokenize entities into wordpiece sequences based on their pronunciations. Despite the positive effects of these techniques, at least in our experiments, they were not sufficient to match hybrid system accuracy on personal named entities.

While most previous work has addressed the personalization of E2E systems with attention-based encoder-decoder (AED) \cite{chan2015listen} or transducer \cite{graves2013speech} architectures, here we begin with a system which uses the Connectionist Temporal Classification (CTC) \cite{graves2006connectionist}. CTC has sparked renewed interest due to its simplicity, competitive accuracy, and amenability to efficient FST decoding (e.g. \cite{pratap2023scaling, zhang2023google, miao2015eesen}.)

Starting from a CTC baseline system, we extend the work of \cite{huang2020class} and present a novel method for generating wordpiece sequences from entity pronunciations.  We show that using this technique in combination with contextual biasing and wordpiece prior normalization, we are able to achieve personal named entity accuracy on par with a competitive hybrid system.

\section{Approach}
\subsection{Baseline decoding architecture}
At the core of our system is a CTC model with wordpiece outputs \cite{kudo2018sentencepiece} and an external word-based LM trained with a large text corpus.  To apply these models to generate ASR hypotheses, our decoder uses a similar strategy to the one described in \cite{miao2015eesen}, with the addition of on-the-fly rescoring \cite{dolfing2001incremental}.  In this strategy, the output of the CTC model serves as the input to a cascade of five FSTs.  The first three of these are $T$ which is built to collapse consecutive matching labels and remove blanks, $L$ which maps collapsed wordpiece sequences to words, and $G_{uni}$, which encodes a word-level unigram LM.  These three FSTs are composed offline to create $TLG_{uni}$.  The next in the cascade, $G^{-1}_{uni}$ serves to remove the effect of $G_{uni}$, and the final FST, $G$, encodes the word-level external LM. Composition with $G^{-1}_{uni}$ and $G$ happens on-the-fly during decoding.

\subsection{Contextual biasing}
\label{subsec:contextual_fst}
As in \cite{zhao2019shallow}, in our application of contextual biasing, we focus specifically on personal named entities.  We implement contextual biasing through a class mechanism similar to the one described in \cite{aleksic2015improved}.  We start by replacing personal named entities with class placeholders in the text used to train our external LM. This allows the external LM to model the likelihood of personal named entity classes occurring given context.  $G$ is constructed from the external LM, with the placeholders preserved.  $TLG_{uni}$ is constructed to include a wordpiece loop that allows any sequence of wordpieces when a class is entered.  At decoding time, the placeholders in $G$ are replaced with contextual bias FSTs which map wordpiece sequences to personal named entities specific to the user.

Construction of contextual bias FSTs follows \cite{huang2020class}. For each class, we create a $G_{c}$ FST from named entities in that class. We use uniform distribution, meaning the probability of each named entity is $1/N_c$, where $N_c$ denotes the total number of named entities in class $c$. We then create $L_c$, an FST to map wordpieces to named entities modeled in $G_c$. These mappings are obtained from both the sentencepiece model and the P2WP model discussed in Section \ref{subsec:p2wp}. Then, we compose, determinize, and minimize $L_c$ and $G_c$ to create a contextual FST.

\subsection{Wordpiece prior normalization}
Our application of contextual biasing can be seen as a way of adapting the external LM to a particular user.  However, as many have observed, the benefit an external LM can provide to an E2E system is often limited by the fact that the E2E model is trained to provide label posteriors and thus contains an ``internal LM'' \cite{mcdermott2019density}.  It has been shown that this internal LM effect can be partly compensated for by constructing an estimate of the internal LM and subtracting its score in log-probability space from the E2E model scores \cite{meng2021internal}.  As in \cite{miao2015eesen}, in this work we use a wordpiece uni-gram trained on the E2E model training data as an estimate of the internal LM.  We subtract the uni-gram wordpiece LM scores from $T$ before it is composed with $L$ and $G$. In effect, this serves to boost the scores of rare wordpieces. To prevent over-boosting, we apply a scale and a clip on the uni-gram LM scores. However, this reduces the costs on non-blank units and encourages insertion errors. To counter that effect, we apply a tune-able cost on the blank unit as well to balance paths through blank or non-blank wordpieces. In fact, we find that even without prior normalization, it is still beneficial to apply a tune-able cost on the blank unit.


\subsection{Pronunciation-driven wordpiece tokenization}
\label{subsec:p2wp}

In some cases, the CTC model scores the wordpiece sequence corresponding to a personal named entity so poorly that wordpiece prior normalization and contextual biasing are not sufficient to compensate. To address these cases and building on \cite{huang2020class}, we developed a technique to generate additional wordpiece representations to be used in contextual biasing.  These representations depend on the pronunciation rather than the orthography of the entity, and they are generated such that they are likely to be given high probabilities by the CTC model given acoustics corresponding to the pronunciation of the entity. 

\begin{figure}[h]
\ninept
\centering
\includegraphics[width=0.4\textwidth]{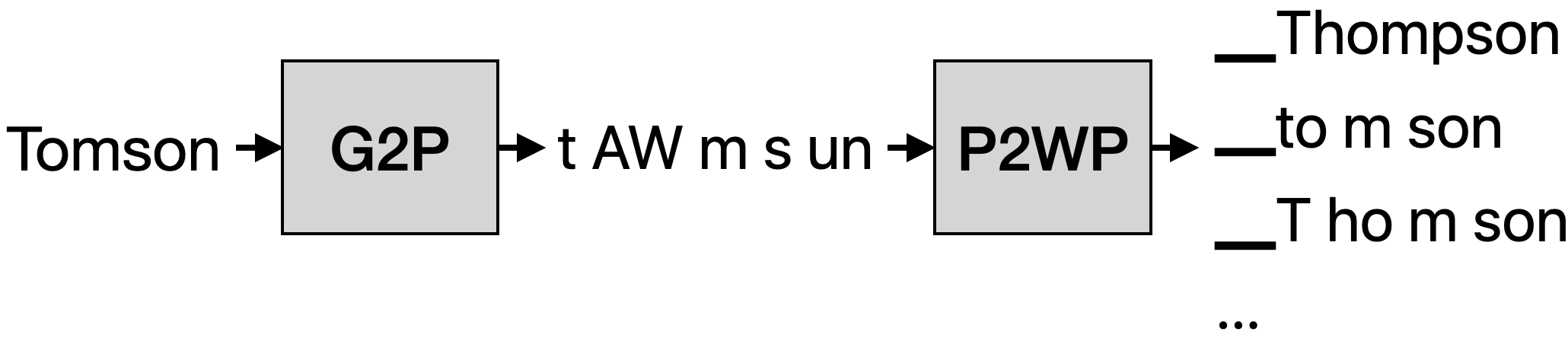}
\caption{An example of pronunciation-driven wordpiece tokenization.}
\label{fig:p2wp}
\end{figure}

As the first step in generating these additional wordpiece representations, entities are decomposed into words from the system lexicon. We obtain a phone sequence for each word from a either a human-curated lexicon or a grapheme-to-phoneme (G2P)
model.  We then apply a phone-to-wordpiece (P2WP) model trained to produce wordpiece sequences that the CTC model will assign high probabilities given an acoustic realization of the input phone sequence.  An example of a word input and wordpiece sequences outputs is shown in Figure \ref{fig:p2wp}.

We train the P2WP model on phone sequence/wordpiece sequence pairs.  We derive the training data from our system lexicon, which provides word/phone sequence pairs.  To obtain the required phone sequence/wordpiece sequence pairs, we apply the wordpiece decomposition used for training the CTC model to the words.  We want the P2WP model to mimic the behavior of the CTC model, so we weight the training data pairs according to the word frequency in the CTC model training data.

We use the Phonetisaurus toolkit \cite{novak2016phonetisaurus} to train and apply the P2WP model.  We configure the training process such that the modeling units are phone sequence/single wordpiece pairs.  The resulting model is a maximum-entropy-trained n-gram on the modeling units.  The model provides the joint probability of a phone sequence, wordpiece sequence, and alignment between the two.  To reduce the size and latency of the resulting model, we perform two kinds of pruning to the n-gram model.  First, we apply entropy pruning \cite{stolcke1998entropy}.  Second, for any phone sequence, we keep only the units with a unigram log probability within a beam of the highest scoring unit for that phone sequence. At runtime, we used the top N-best scoring representations according to the P2WP model for each word in each entity.

\section{Related work}
Our work is most related to \cite{huang2020class}.  In that work, as in ours, contextual biasing FSTs are constructed from personal entities and used via a class mechanism.  In addition, that work describes a method to create additional wordpiece representations based on entity pronunciations.   However, our work advances that work in important ways, specifically in the method to create pronunciation-based wordpiece representations.  The method described there first maps phoneme sequences to word sequences, then uses the baseline wordpiece tokenizer to obtain wordpiece representations.  The resulting system is thus constrained to produce wordpiece sequences corresponding to full word sequences.  As we show in Section \ref{subsec:results}, our approach has a smaller memory footprint and achieves better accuracy on personal named entities.  That work also differs from ours in that the E2E system used in that work uses an AED architecture, while our work is to personalize a CTC-based system.  Relatedly, that work does not use an FST-based graph decoder, and thus the class mechanism is implemented differently.

Another way to generate additional wordpiece representations for rare words is described in \cite{le2020g2g}. In that work, additional representations are generated by first running speech synthesis from word orthographies, then running a lexicon-free decode on the resulting audio to get character-level representations. Compared to our approach, this approach has the disadvantage that it requires running speech synthesis and decoding. The resulting representations may also be affected by the particular speech synthesis engine being used.

Neural contextual biasing is another thread of related work. These methods have shown promising results and are appealingly elegant but have not yet been proven to work with large entity lists and may have latency implications \cite{pundak2018deep}.

\section{Experiments}
\subsection{Models and test sets}
We train a hybrid AM and a CTC E2E model following \cite{DBLP:conf/interspeech/YaoWWZYYPCXL21} and \cite{huang2020sndcnn}. We use the model structure and parameters recommended by those works, except that the output size of the CTC model is adjusted to fit our in-house wordpiece list. The CTC model is powered by a 12-layer Conformer \cite{gulati2020conformer} encoder. Though the CTC model is trained with a multitask objective of CTC and AED, we use as our baseline a streaming setup without AED rescoring. We use a 240ms decoding window. It is worth noting that for American English and German, we use wordpieces tokenized by a sentencepiece model. For Chinese Mandarin, we use character tokenization. In the rest of the paper, we still call the grapheme units in Chinese Mandarin wordpieces unless otherwise noted. We use a 4-gram language model trained on multi-domain text corpora in both the hybrid and the CTC systems. The P2WP models are 5-grams, entropy-pruned with a threshold of 1e-7 for \texttt{en} and \texttt{de} and 1e-8 for \texttt{zh}. P2WP model units are pruned using a beam of 3.

We carry out experiments on a voice assistant task in three languages: American English, Chinese Mandarin, and German. The test sets are anonymized, human-annotated utterances randomly sampled from virtual assistant queries. Though our solution naturally supports multiple classes, we only use the contact class in the experiments below, as it is the dominant use case. We further divide the test sets into subsets A and B, which contains and does not contain contact named entities, respectively, to better understand our solution's accuracy impact. The total number of contact named entities in the test sets is also counted. Table \ref{tab:testset} shows the statistics of the test sets. 

\begin{table}[h]
\ninept
    \centering
    \begin{tabular}{lllll}
        \hline
        \multirow{2}{*}{Language} & \multicolumn{3}{c}{\#Utterances} & \multirow{2}{*}{\#Contacts} \\
        \cline{2-4}
         & All & Subset A & Subset B & \\
         \hline
         en & 49.4k & 18.4k & 31.0k & 18.5k \\
         zh & 33.5k & 5.2k & 28.3k & 5.2k \\
         de & 28.1k & 5.5k & 22.6k & 5.6k \\
         \hline
    \end{tabular}
    \caption{Statistics of the test sets. Subset A and B are subsets containing and not containing contacts, respectively.}
    \label{tab:testset}
\end{table}

\subsection{Results and analysis}
\label{subsec:results}
\begin{table*}[h]
\ninept
    \centering
    \begin{tabular}{l|cccc|cccc|cccc}
        \hline
\multirow{3}{*}{System}       & \multicolumn{4}{c|}{en} & \multicolumn{4}{c|}{zh} & \multicolumn{4}{c}{de} \\
        \cline{2-13}
         & \multicolumn{3}{c|}{WER} & \multirow{2}{*}{CEER} & \multicolumn{3}{c|}{WER} & \multirow{2}{*}{CEER} & \multicolumn{3}{c|}{WER} & \multirow{2}{*}{CEER} \\
         \cline{2-4} \cline{6-8} \cline{10-12}
        & All & A & \multicolumn{1}{c|}{B} & & All & A & \multicolumn{1}{c|}{B} & & All & A & \multicolumn{1}{c|}{B} & \\
        \hline
        Hybrid & 3.88 & 2.40 & 4.44 & \textbf{3.48} & 5.33 & 1.78 & 5.93 & 3.51 & 5.40 & 2.78 & 5.94 & 4.20 \\
        \hline
        CTC+4-gram & 6.95 & 13.80 & 4.33 & 36.3 & 6.51 & 13.01 & 5.42 & 49.23 & 6.89 & 13.52 & 5.53 & 40.38 \\
        \hspace{3mm}+Personalized & 4.13 & 3.65 & 4.31 & 7.18 & 4.87 & 2.10 & \textbf{5.34} & 4.81 & 5.11 & 3.56 & 5.43 & 6.76 \\
        \hspace{6mm} +P2WP=1 & 3.92 & 2.90 & 4.31 & 4.93 & \textbf{4.86} & 1.88 & 5.35 & 4.01 & 5.04 & 3.12 & 5.44 & 5.3 \\
        \hspace{6mm} +P2WP=4 & 3.86 & 2.63 & 4.33 & 4.25 & \textbf{4.86} & 1.75 & 5.38 & 3.59 & 5.03 & 2.93 & 5.46 & 4.76 \\
        \hspace{9mm} +WP norm & \textbf{3.73} & \textbf{2.29} & \textbf{4.28} & 3.67 & 4.92 & \textbf{1.63} & 5.47 & \textbf{3.44} & \textbf{4.93} & \textbf{2.69} & \textbf{5.29} & \textbf{4.04} \\
      \hline
    \end{tabular}
    \caption{Overall WER, WER on the subset containing / not containing contacts, and contact entity error rate. The results show that P2WP and wordpiece normalization consistently improve personal named entity accuracy across languages.}
    \label{tab:results}
\end{table*}

First, we compare the pronunciation-driven tokenization approach presented in \cite{huang2020class} with our P2WP model, with results for American English shown in Table \ref{tab:lg_vs_p2wp}. We follow their work and build the LG FST used in their method with the same pronunciation lexicon used to build the P2WP FST. Then, we build a personalized CTC system and use either the LG FST or the P2WP FST to provide additional pronunciation-driven wordpiece sequences. To get a word's pronunciation-driven wordpieces, we first look it up in an existing pronunciation lexicon, and if not found, we run this word through G2P to get its 4-best pronunciations. We then feed all pronunciations to the LG FST or the P2WP model to get the 10-best wordpiece sequences for each.  Wordpiece prior normalization is also used with scale=0.8, clip=20, and blank cost being -3, all tuned on a dev set. The systems are benchmarked on overall WER and contact named entity error rate, denoted as \texttt{WER} and \texttt{CEER}, respectively. \texttt{CEER} is defined as the percentage of misrecognized contacts. The results show that despite being over 10x smaller than the LG FST measured by the number of FST arcs, the P2WP model yields better personal named entity accuracy, a 13.3\% relative reduction. While P2WP is also faster than the LG FST in our experience, we do not compare their speed as a beam search algorithm is provided by Phonetisaurus for P2WP. Meanwhile for the LG FST, we use standard FST operations without pruning.

\begin{table}[h]
\ninept
    \centering
    \begin{tabular}{l|llll}
        \hline
        Model & \#Arcs & WER & CEER \\
        \hline
        LG FST \cite{huang2020class} & 5.44M & 3.77 & 4.30 \\
        P2WP & 0.53M & 3.70 & 3.73 \\
        \hline
    \end{tabular}
    \caption{Comparison between P2WP and the LG FST presented in \cite{huang2020class}.}
    \label{tab:lg_vs_p2wp}
\end{table}

We then conduct experiments in the aforementioned three languages to better understand the accuracy impact of  the P2WP model and wordpiece prior normalization. Other than \texttt{WER} and \texttt{CEER}, we also look at WER on subset A and B, denoted as \texttt{WER A} and \texttt{WER B} respectively. We use 1-best and 4-best P2WP outputs for each input pronunciation, denoted as \texttt{P2WP=1} and \texttt{P2WP=4}. The costs on the blank unit are tuned for all experiments on a dev set. For systems with or without wordpiece prior normalization (with scale=0.8 and clip=20, tuned on a dev set), optimal costs of -3 and 3 are found, respectively. It is worth noting that models in Table \ref{tab:results} and \ref{tab:lg_vs_p2wp} are trained on different data and are hence incomparable.

The results in Table \ref{tab:results} show that our class-based implementation of contextual biasing brings a huge reduction on personal named entity error rate over no personalization. However, there is still a massive gap between the CTC system and the strong hybrid baseline.  Including additional wordpiece tokenizations generated using the P2WP model further reduces the gap, leading to up to 31.3\% relative reduction on \texttt{CEER} (en, 7.18 to 4.93), and up to 20.5\% relative WER reduction on the contact name-rich subset (en, \texttt{WER A}, 3.65 to 2.9). A larger P2WP N-best list (1 to 4) expands the \texttt{CEER} reduction to up to 40.8\% (en, 7.18 to 4.25), and \texttt{WER A} reduction to up to 27.9\% (en, 3.65 to 2.63). Despite huge improvement on personal named entity recognition, using the P2WP model only marginally degrades (by up to 0.75\%, zh, 5.34 to 5.38) the non-contact subset B. Interestingly, we find that applying the P2WP model decreases WER more on American English than Chinese Mandarin and German. This is likely because German and Chinese Mandarin have more predictable grapheme-phoneme relations, so it is easier for the E2E model to predict a rare word given the speech. 

\begin{figure}[h]
\ninept
\centering
\includegraphics[width=0.35\textwidth]{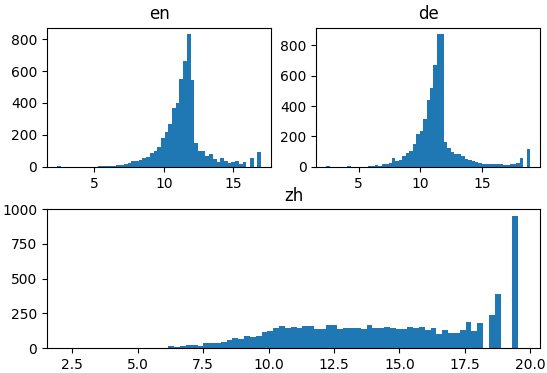}
\caption{Histogram of wordpiece uni-gram LM costs. X-axis represents LM costs and Y-axis represents wordpiece counts. \texttt{en} and \texttt{de} have similar distribution while \texttt{zh} has a much flatter distribution with many infrequent wordpieces.}
\label{fig:hist}
\end{figure}

We further apply wordpiece prior normalization, which consistently improves all four metrics in American English and German, leading to 48.9\% reduction over the baseline personalized CTC in English. However, in Chinese Mandarin, wordpiece normalization only marginally improves \texttt{CEER}, but hurts \texttt{WER B} and the overall WER. To better understand the problem, we draw a histogram of the uni-gram LM costs of wordpieces for all three languages, shown in Figure \ref{fig:hist}. We can see that English and German have a similar distribution peaking at around 12. The uni-gram LM cost distribution in Chinese Mandarin is flatter likely because we use character tokenization rather than sentencepiece. Also, the histogram peaks at around 20, causing many overly infrequent wordpieces (characters) to be boosted, leading to an overall accuracy degradation.

\section{Conclusion}

Building on the work of \cite{huang2020class}, we present a novel method for generating wordpiece sequences from entity pronunciations.  We show that by using this method to generate additional personal entity representations for contextual biasing and applying wordpiece prior normalization, we are able to match the accuracy of high-quality hybrid systems on personal named entity recognition. 
\bibliographystyle{IEEEbib}
{\ninept\bibliography{refs}}

\end{document}